\documentclass[sigconf]{acmart}

\usepackage{enumitem}
\usepackage{booktabs, makecell, multirow}
\usepackage{graphicx}
\usepackage{subcaption}
\usepackage{float}
\usepackage{amsmath}
\usepackage{xspace}
\usepackage{url}
\usepackage{tabularx, multirow, multicol}
\usepackage[ruled,vlined]{algorithm2e}
\usepackage{pifont}
\usepackage{color, xcolor}
\AtBeginDocument{%
  }

\copyrightyear{2023} 
\acmYear{2023} 
\setcopyright{acmlicensed}\acmConference[SEC '23]{The Eighth ACM/IEEE Symposium on Edge Computing}{December 6--9, 2023}{Wilmington, DE, USA}
\acmBooktitle{The Eighth ACM/IEEE Symposium on Edge Computing (SEC '23), December 6--9, 2023, Wilmington, DE, USA}
\acmPrice{15.00}
\acmDOI{10.1145/3583740.3626617}
\acmISBN{979-8-4007-0123-8/23/12}

\newcommand{\system}{\textit{SODA}\xspace}
\newcommand{\cmark}{\ding{51}}
\newcommand{\xmark}{\textcolor{lightgray}{\ding{55}}}

\begin{document}

%%
%% The "title" command has an optional parameter,
%% allowing the author to define a "short title" to be used in page headers.
% \title[Protecting Proprietary Information During On-Device Deployment]{Protecting Proprietary Information During On-Device Deployment of Machine Learning Models}
\title{SODA: Protecting Proprietary Information in On-Device Machine Learning Models}

%%
%% The "author" command and its associated commands are used to define
%% the authors and their affiliations.
%% Of note is the shared affiliation of the first two authors, and the
%% "authornote" and "authornotemark" commands
%% used to denote shared contribution to the research.
\author{Akanksha Atrey$^1$, Ritwik Sinha$^2$, Saayan Mitra$^2$, Prashant Shenoy$^1$}
\affiliation{
\institution{$^1$University of Massachusetts Amherst, $^2$Adobe Research}
\country{}}
\email{aatrey@cs.umass.edu, {risinha,smitra}@adobe.com, shenoy@cs.umass.edu}
% \author{Akanksha Atrey}
% \affiliation{%
%   \institution{The Kumquat Consortium}
%   \city{New York}
%   \country{USA}}
% \email{jpkumquat@consortium.net}

% \author{Ritwik Sinha}
% \affiliation{%
%   \institution{The Kumquat Consortium}
%   \city{New York}
%   \country{USA}}
% \email{jpkumquat@consortium.net}

% \author{Saayan Mitra}
% \affiliation{%
%   \institution{The Kumquat Consortium}
%   \city{New York}
%   \country{USA}}
% \email{jpkumquat@consortium.net}

% \author{Prashant Shenoy}
% \affiliation{%
%   \institution{The Kumquat Consortium}
%   \city{New York}
%   \country{USA}}
% \email{jpkumquat@consortium.net}

%%
%% By default, the full list of authors will be used in the page
%% headers. Often, this list is too long, and will overlap
%% other information printed in the page headers. This command allows
%% the author to define a more concise list
%% of authors' names for this purpose.
\renewcommand{\shortauthors}{Atrey et al.}

%%
%% The abstract is a short summary of the work to be presented in the
%% article.
\begin{abstract}
   The growth of low-end hardware has led to a proliferation of machine learning-based services in edge applications. These applications gather contextual information about users and provide some services, such as personalized offers, through a machine learning (ML) model. A growing practice has been to deploy such ML models on the user's device to reduce latency, maintain user privacy, and minimize continuous reliance on a centralized source. However, deploying ML models on the user's edge device can leak proprietary information about the service provider. In this work, we investigate on-device ML models that are used to provide mobile services and demonstrate how simple attacks can leak proprietary information of the service provider. We show that different adversaries can easily exploit such models to maximize their profit and accomplish content theft. Motivated by the need to thwart such attacks, we present an end-to-end framework, \system, for deploying and serving on edge devices while defending against adversarial usage. Our results demonstrate that \system can detect adversarial usage with 89\% accuracy in less than 50 queries with minimal impact on service performance, latency, and storage.
\end{abstract}

%%
%% The code below is generated by the tool at http://dl.acm.org/ccs.cfm.
%% Please copy and paste the code instead of the example below.
%%


%%
%% Keywords. The author(s) should pick words that accurately describe
%% the work being presented. Separate the keywords with commas.

%%
%% This command processes the author and affiliation and title
%% information and builds the first part of the formatted document.
\maketitle

\section{Introduction}
The ubiquity of machine learning (ML) models in distributed applications such as fitness tracking, entertainment recommendations, virtual personal assistance, and social media services has changed the way humans interact with their devices. This proliferation has led to consumers being more proactive and conscious about their choices, including about what data leaves their edge devices \cite{dobelt2015consumers} and the need for faster response times \cite{arapakis2014impact}. This implies that ML-based predictions and recommendations cannot be conducted in a centralized manner and require them to be served closer to where the consumer is. For instance, consider a personalization model that takes as input the context of the user and recommends entertainment choices. With the advancement of low-end hardware technologies \cite{apple2017neural, intelmovidius, nvidia2019EGX}, this inference can be conducted on the consumer's device (i.e., mobile phone).

%advantages of on-device models
Conducting inference on the end user's device is advantageous in many aspects \cite{Advantag49:online}. First, serving on the device reduces the service latency seen by the end user. Second, keeping the model on the device preserves user privacy as all the user data processing happens on the device. Third, on-device models are capable of running offline without continuous network connectivity to the centralized server. Lastly, running inference on the device reduces cloud compute cycles, consequently reducing costs for the service provider. 
However, on-device ML models are prone to exploitation since they lie outside the natural security perimeter of the service provider cloud that could detect, track, and protect against adversarial actions. We examine the privacy of on-device models from a service provider's point of view in this paper. 

%Privacy threat for on-device models
Previous works have proposed attacks to steal ML models in ML-as-a-service applications by training substitute models \cite{papernot2017practical, tramer2016stealing, oh2019towards, juuti2019prada}. This approach considers the model as intellectual property, necessitating considerable effort to train substitute models that exhibit comparable performance.
However, once a model is deployed on the device, the primary concern shifts from model access to safeguarding the proprietary information embedded within the model. We argue that extracting proprietary information from models requires substantially fewer resources than model stealing.

\begin{figure}
    \centering
    \begin{subfigure}[b]{0.22\textwidth}
    \includegraphics[width=\textwidth]{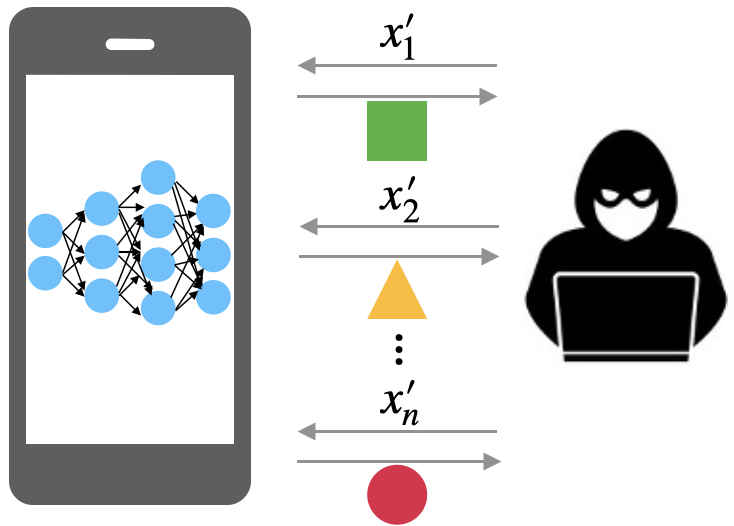}
    \caption{}
    \label{fig:adv_goal1}
    \end{subfigure}
    \hfill
    \begin{subfigure}[b]{0.22\textwidth}
    \includegraphics[width=\textwidth]{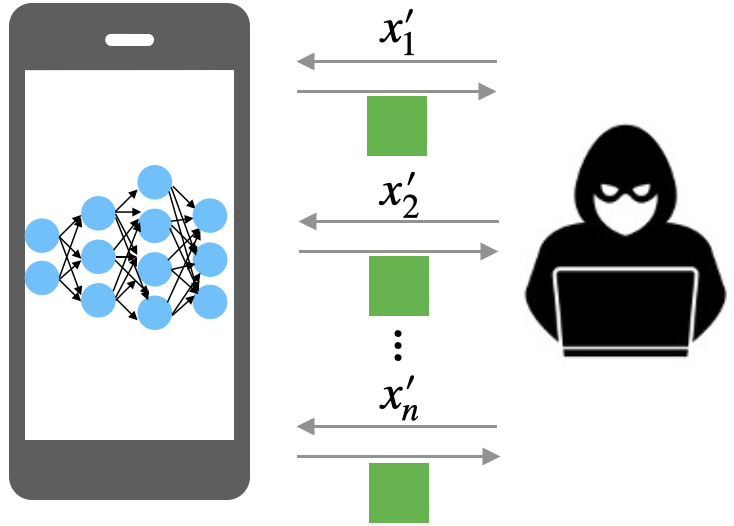}
    \caption{}
    \label{fig:adv_goal2}
    \end{subfigure}
    \caption{Examples of leaking proprietary information via on-device ML models where $x'_i$ represents adversarial queries and shapes represent model output (e.g., service): (a) exploiting output diversity, and (b) exploiting decision boundaries of a particular service type (e.g., class).}
    \label{fig:adversary_goals}
    \vspace{-1em}
\end{figure}

In this work, we define the adversarial goal to be to steal proprietary information embedded in an ML model that is located on the device. For example, consider a bank application that produces personalized credit card offers based on users’ contextual information. Assume a ML model is deployed on the edge for this purpose. Here the distribution and range of offers available along with `who is recommended what' is proprietary information. An adversary can learn what sets of inputs give them the most profitable output by learning the probability distribution of the potential offers as shown in Figure \ref{fig:adv_goal1}. Alternatively, consider an image authentication service that accepts or rejects images on a social media platform based on their authenticity. Here the criteria or rules embedded in the model are proprietary information. An adversary can learn the rules by investigating the model's most salient features and distort the input in minor ways to authenticate disallowed images as shown in Figure \ref{fig:adv_goal2}. 

In both the examples above, a common theme is to exploit the hidden representations or input-to-output mappings that are embedded in the ML model. Often these are rules or criteria set by the service provider and are considered proprietary information. Having unauthorized access to this information is disadvantageous to the brand. Firstly, this information can be published or sold to competitors, coupon sites, or price aggregators which can affect the service provider's business. Secondly, since such systems often have a feedback loop with new streams of data being used for model updates, the adversary can poison or bias the future versions of the model with their high volume of atypical queries.

In this paper, we examine these privacy issues in on-device models from the service provider's point of view. Motivated by the need to protect the service provider's proprietary information, we propose Secure On-Device Application (\system), an end-to-end system for deploying and serving on device. \system defends the proprietary information in on-device ML models using an autoencoder-based approach that captures adversarial usage across time. In the design and implementation of \system, we make the following contributions:

\begin{enumerate}[label=\textbf{C\arabic*}, topsep=2pt]
    \item We develop a taxonomy of on-device models by examining models used in distributed services (e.g., web or mobile applications).
    \item We demonstrate how simple privacy attacks can leak proprietary information contained in on-device models with differing levels of threats. We group these threats into white-box and black-box attacks.
    % \item We evaluate the efficacy of the privacy attacks on three real-world datasets and empirically analyze the characteristics that make on-device models more susceptible to such attacks. Our results demonstrate the easability with which such attacks can leak sensitive proprietary information of the service provider.
    \item We propose a robust system, \system, to defend against leakage of proprietary information from on-device models. The proposed solution protects against differing levels of threats ensuring its generalizability and adaptability.
    \item Our empirical evaluation on two widely used datasets demonstrates that \system can detect adversarial usage with 89\% accuracy in less than 50 queries with a minimal increase in latency.
\end{enumerate}

\section{Background}
In this section, we present background on distributed services and the usage of ML in such services.

\subsection{Distributed Services}
This work focuses on distributed services whose service components are distributed across edge devices and cloud. Traditional services host the front-end components on the edge device and back-end components on the cloud. Modern day services collect contextual data about the user and their environment to provide data-dependent services. These services can be categorized into three: one-time, occasional, and real-time. Examples of one-time services include financial offers or life insurance risk assessments. Occasional services range from health or smart home applications which are used on a per-need basis. Finally, real-time services involve mapping applications and fitness trackers. We focus on one-time and occasional services. Real-time services offer a different set of challenges due to their continuous usage.

\subsection{Machine Learning in Distributed Services}
Much of the data-intensive computation in distributed services is aided by data mining and ML. For example, consider ride-sharing applications which use ML for next location prediction. An ML model is typically trained on the cloud using data from many users and each service query is served through a direct call to the model's API. 

With the development of low-end hardware (e.g., Apple's Neural Engine \cite{apple2017neural} and Intel's Movidius \cite{intelmovidius}), more computation is moving to the end user's device. That is, models can now be trained or fine-tuned on the edge device itself using user-specific data and can be served on the device for faster inference \cite{yoon2017efficient, atrey2021preserving}. In this work, we focus specifically on models that are deployed on edge devices for inference regardless of where they are trained. We assume the training process is kept distinct from the end user.

\subsection{Privacy Attacks in Machine Learning}
Data privacy has become an increasing concern with the proliferation of ML. Prior works have proposed attacks which leak information about sensitive features in the training data through model inversion \cite{fredrikson2015model}, membership of data samples through membership inference \cite{shokri2017membership}, and embedded global patterns in the data through property inference \cite{ganju2018property}. Closer to our work, model extraction attacks have been proposed to leak information about the model itself to create copies of models locally \cite{tramer2016stealing}. This requires recursively querying the target model to build a substitute data set for training a shadow model. These types of attacks are especially harmful in ML-as-a-service applications where commercially valuable models are allowed to be used on a pay-per-query basis. While these attacks apply in on-device deployment settings, training shadow models require a sufficiently large number of queries. We instead focus specifically on the privacy of the proprietary information of the service provider embedded in ML models, which can be leaked in much fewer queries in comparison. To the best of our knowledge, no prior work considers this facet of proprietary information leakage in on-device models.

\begin{table*}[t]
\small
\centering
\caption{Taxonomy of white-box (WB) and black-box (BB) ML models on the device with the components of the ML model accessible by a user.}
\begin{tabular}{c|cc|cc|ccc}
\hline
 \textbf{Model} & \multicolumn{2}{c}{\textbf{Feature Space}}    & \multicolumn{2}{c}{\textbf{Output Space}} & \multicolumn{3}{c}{\textbf{Internals}} \\
\textbf{Type}& All & Model Input & Model Output & Output Probabilities & Architecture & Parameters   & Representations \\
\hline
WB     & \cmark   & \cmark  & \cmark  & \cmark  & \cmark & \cmark & \xmark \\
BB  & \cmark   & \xmark      & \cmark & \xmark      & \xmark  & \xmark  & \xmark   \\
\hline
\end{tabular}
\label{tab:ondevice_taxonomy}
\end{table*}

\subsection{Privacy Preserving Model Serving}
With the emergence of novel attacks targeting model deployment and serving, limiting queries or following black-box deployment methods are naive strategies for inference privacy. However, such methods are not always feasible, especially in distributed applications where ensuring benign users are provided appropriate service is important. More sophisticated methods have been proposed to preserve inference privacy with the goal of protecting private information in the training data. These methods rely on traditional approaches such as differential privacy, homomorphic encryption, and information theoretic privacy \cite{papernot2016distillation, liu2017oblivious, papernot2017semi, mireshghallah2020shredder}. 
% Furthermore, traditional methods such as differential privacy and homomorphic encryption are primarily used to secure data samples or trained models. 
However, these solutions do not consider privacy from the \emph{service provider's standpoint} since high-level representations can still be leaked via continuous querying. 

% minionn -- \cite{liu2017oblivious} -- convert neural network to oblivious where client does not have access to user data and user does not have access to anything except the output. Uses additive homomorphic encryption (AHE) in a
% preprocessing step.
% pate -- \cite{papernot2017semi} -- Private Aggregation of Teacher Ensembles, ensemble of models trained on sensitive data and student models are trained to predict output chosen by noisy voting by teacher models.
% knowledge distillation -- \cite{papernot2016distillation} -- models trained with defensive distillation to make more robust to adversarial perturbations

\section{Privacy Implications of On-Device Models}
In this section, we set up the privacy problem in models that are deployed on edge devices for inference. We first present a taxonomy of on-device models and then demonstrate how such models can be attacked to leak proprietary information about the service provider.

\subsection{Taxonomy of On-Device Models}

An ML model can be divided into three spaces: feature space, prediction space, and internals. The feature space includes inputs to the ML system, including features that are collected by the application but not employed during the training of the model. The prediction space includes model predictions and the probabilities associated with those predictions. Finally, the internals space consists of the model architecture, parameters (e.g., weights), and hidden representations.

%See: https://ubuntu.com/blog/guide-to-ml-model-serving
The deployment of ML models on the device can be categorized into white-box deployment and black-box deployment as summarized in Table \ref{tab:ondevice_taxonomy}. White-box deployment provides users transparency and access to all components in the feature and output spaces, and model parameters. This is equivalent to deploying a serialized version of the model such that it is programmatically accessible by any modern framework. Popular serialization methods include Tensorflow SavedModel, Open Neural Network Exchange (ONNX), Predictive Model Markup Language (PMML), and TorchScript. Serialized models can be deployed as an external file on a web or mobile application, and used as a library by the application. In such settings, the model is accessible via the browser's inspect element, sophisticated web scraping methods, or by accessing the application bundle\footnote{Note, accessing the application bundle requires appropriate access controls.}.

Black-box deployment provides a layer of security by only revealing the inputs collected by the application and the model output. This is equivalent to deploying a model in a mobile application by embedding it into the application's binary interface. Alternatively, deploying encrypted serialized model files, adding access controls on serialized model files, or adding secure boot and firmware protections are also black-box deployment methods. Unless the adversary has memory access, the model components are much harder to access in these situations. 

\subsection{Threat Model}
\label{sec:threat-model}
We describe the entities of the threat model as follows.

\vspace{0.5em}
\noindent
\textbf{Service Provider.} We consider a service provider $\mathcal{S}$ who is responsible for providing an arbitrary service. $\mathcal{S}$ employs ML-based predictions to aid its service by training a general, multi-user model $M: X \to Y$ where $X$ is contextual data and $Y$ represents the set of possible service categories, such as credit card offers. We assume that $M$ is a $C$-class model trained on the cloud, where $C$ represents the number of distinct classes in $Y$ (i.e., the cardinality of the response), and the training process is kept distinct from other entities in the threat model. This model is then deployed on the end user's device for serving. We assume $Y$ and the representations learned by $M$ are proprietary information. 

\vspace{0.5em}
\noindent
\textbf{User.} We consider a user $\mathcal{U}$ who uses the service provided by $\mathcal{S}$. We assume $\mathcal{U}$ stores the deployed model $M$ on their edge device and allows the service to use their contextual data, $X_\mathcal{U}$. $\mathcal{U}$ is expected to accept the service (e.g., $M(X_\mathcal{U}) \to Y$) provided by $\mathcal{S}$ in an honest manner.

\vspace{0.5em}
\noindent
\textbf{Adversary.} We consider an adversary $\mathcal{A}$ who uses the service provided by $\mathcal{S}$ in order to extract proprietary information about $\mathcal{S}$. Similar to $\mathcal{U}$, $\mathcal{A}$ stores the deployed model $M$ on their device and is expected to accept the service (e.g., $M(X_\mathcal{A}) \to Y$). As a baseline, the adversary has access to an organic query $x_\mathcal{A}$. We assume $\mathcal{A}$ to be a curious adversary who attempts to learn $M(X'_\mathcal{A}) \to Y$ with two alternative goals:
\begin{enumerate}[label=\textbf{A-\arabic*}, itemsep=1pt]
    \item Identify all $y_c$ where $c \in \{1, 2,...,C\}$ such that $M(X'_\mathcal{A}) \to Y$ (i.e., exploiting output diversity).
    % (i.e., exploiting $X_\mathcal{A}$ to recover the output space).
    \item Identify variances $x'_\mathcal{A} \sim x_\mathcal{A}$ such that $M(X'_\mathcal{A}) \to y_c$ (i.e., exploiting decision boundary of class $c$).
\end{enumerate}
We explore $\mathcal{A}$'s ability to query the model in both white-box and black-box scenarios. Since $M$ is stored on their device and is accessible in an offline fashion, $\mathcal{A}$ is not bounded by query limits.
%Finally, $\mathcal{A}$ can download the model in white-box scenarios and only query in black-box scenarios.
We further assume the adversary lacks the capability to inspect the memory of a running program and disassemble RAM to extract executable code or the model executable.

\begin{table}
\centering
\small
\caption{Performance (\%) of querying attacks across datasets and models. Attack A-1 demonstrates the percentage of prediction space recovered using randomly generated queries, and attack A-2 demonstrates the accuracy of exploiting decision boundaries across classes via random perturbations.}
\begin{tabular}{c|c|c|c} 
\hline
\multicolumn{2}{c}{} & \textbf{Attack A-1} & \textbf{Attack A-2} \\
\hline
\multirow{3}{*}{\textbf{HAR}} & \textbf{Random Forest} & 66.33 & 96.18 \\
                             & \textbf{Logistic Regression} & 100.00 & 99.85 \\
                             & \textbf{Deep Neural Network} & 100.00 & 99.72 \\
\hline
\multirow{3}{*}{\textbf{MNIST}} & \textbf{Random Forest} & 29.50 & 98.67 \\
                                 & \textbf{Logistic Regression} & 100.00 & 99.73 \\
                                 & \textbf{Deep Neural Network} & 90.50 & 100.00 \\
\hline
\end{tabular}
\label{tab:attack_results}
\end{table}

\begin{figure*}[t!]
    \centering
    \begin{subfigure}[b]{0.32\textwidth}
    \includegraphics[width=\textwidth]{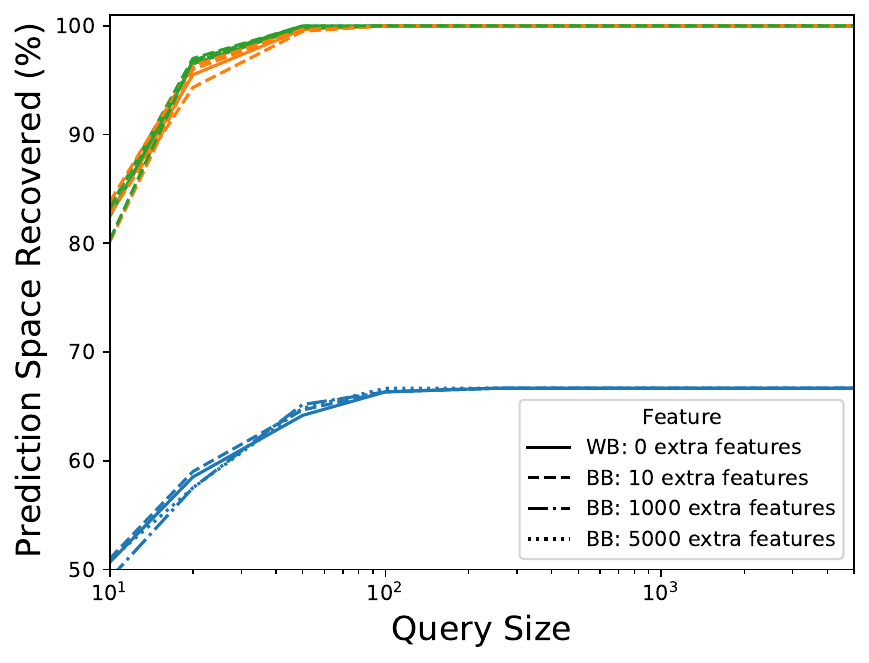}
    \caption{}
    \label{fig:class_attacks1}
    \end{subfigure}
    \begin{subfigure}[b]{0.32\textwidth}
    \includegraphics[width=\textwidth]{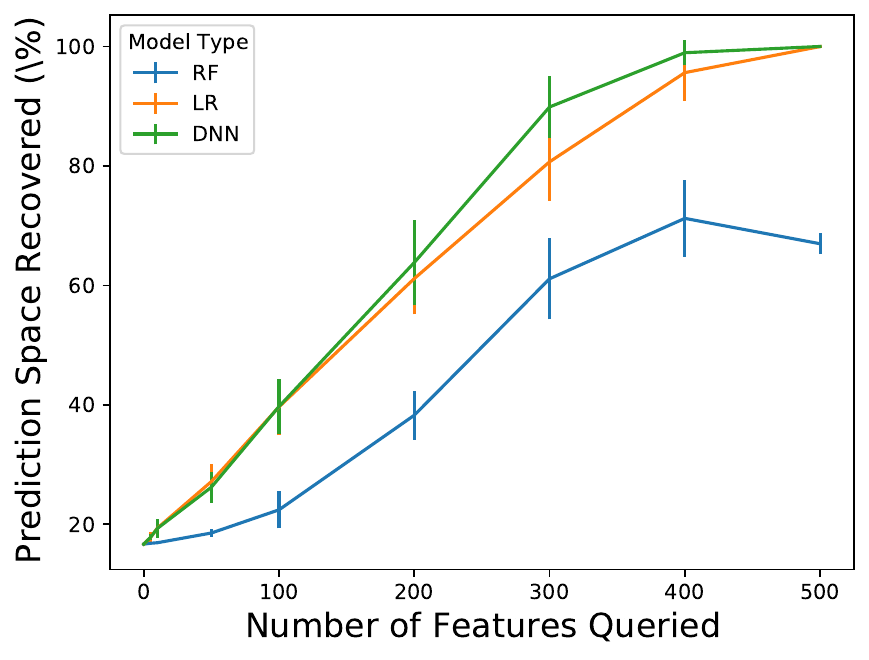}
    \caption{}
    \label{fig:class_attacks2}
    \end{subfigure}
    \begin{subfigure}[b]{0.32\textwidth}
    \includegraphics[width=\textwidth]{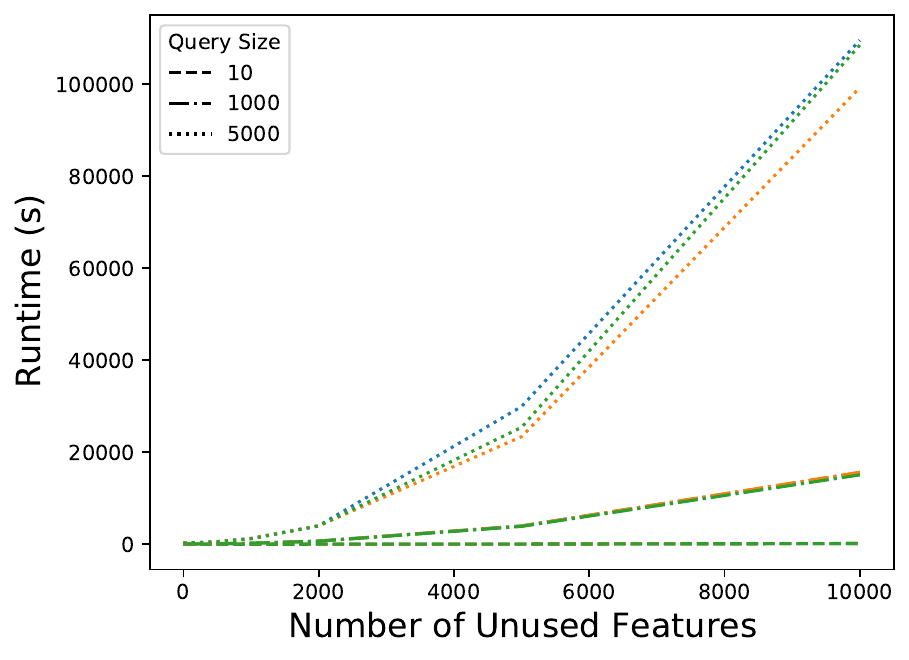}
    \caption{}
    \label{fig:class_attacks3}
    \end{subfigure}
    \caption{Results of random querying attacks on decision trees (DT), multi-class logistic regression models (LR) and deep neural networks (DNN) in white-box (WB) and black-box (BB) environments: (a) impact of query size on the classes recovered; (b) impact of the number of features queried among the model input on the classes recovered; and (c) impact of the number of unused features on runtime in the BB environment. The legend in (b) applies to all figures.}
    \label{fig:class_attacks}
\end{figure*}

\subsection{Exploiting Proprietary Information Through Querying Attacks}
\label{sec:attack}
While prior works have proposed attacks to steal ML models by training substitute models, the order of queries required to build a substitute model is on the scale of 1000s \cite{papernot2017practical, juuti2019prada, orekondy2019knockoff}. In this work, we demonstrate that only a limited number of queries are needed to leak proprietary information in on-device ML models. Specifically, we show querying attacks starting with a single seed query are sufficient to leak information about input-to-output mappings with as little as 50 queries.
While more sophisticated methods may be used for querying, our goal is to demonstrate how simple attacks can leak varying levels of proprietary information. 
The simplicity of the attack is aided by the nature of on-device models which are often accessible in an offline fashion and not restricted in usage. 

\subsubsection{Experimental Setup}
\label{sec:attack-setup}
To test the efficacy of querying attacks, we use two popular datasets: (1) UCI's human activity recognition (HAR) dataset \cite{anguita2013public}, and (2) MNIST digits recognition dataset \cite{mnist:online}. The HAR dataset consists of 561 smartphone sensor features from 30 users while performing six activities (walking, walking upstairs, walking downstairs, sitting, standing, and laying). All features are normalized and bounded to a [-1, 1] scale. The training data consists of a random partition of 70\% of the volunteers, while the test data comprises the remaining 30\% of the participants. The MNIST dataset contains 70,000 28x28 grayscale images of handwritten digits from 0 to 9. Each pixel in the image is represented by a number from 0 to 255. We normalize the features to a [-1, 1] scale and transform the images into one-dimensional vectors of 784 features. We use 60,000 images for training and the remaining 10,000 images as test data.

We train random forests (RF), multi-class logistic regression models (LR) and deep neural networks (DNN) for the human activity recognition and digits recognition tasks. Optimal hyperparameters are chosen for each model by performing randomized search on 3-fold cross validation. The resulting recognition performance is 96\%, 93\%, and 94\% for the HAR dataset respectively, and 97\%, 92\%, and 96\% for the MNIST dataset respectively.

The attacks are conducted using 100 randomly selected organic seed queries from the test data. All results are aggregated across these 100 adversaries.

\begin{table}
\centering
\small
\caption{Runtimes of random query attacks to recover maximum percentage of the prediction space before plateauing (A-2).}
\begin{tabular}{c|c|c} 
\hline
\textbf{Model Type} & \textbf{Attack Type} & \textbf{Runtime (s)} \\
\hline
\multirow{2}{*}{RF} & White-box & 0.2801 \\
 & Black-box & 54.2049 \\ %rf : 54.2049 \\
\hline
\multirow{2}{*}{LR} & White-box & 0.0951 \\
 & Black-box & 18.9895 \\
\hline
\multirow{2}{*}{DNN} & White-box & 0.0929 \\
 & Black-box & 18.4184 \\
\hline
\end{tabular}
\label{tab:class_attack_runtime}
\end{table}

\begin{figure*}[t!]
    \centering
    \begin{subfigure}[b]{0.32\textwidth}
    \includegraphics[width=\textwidth]{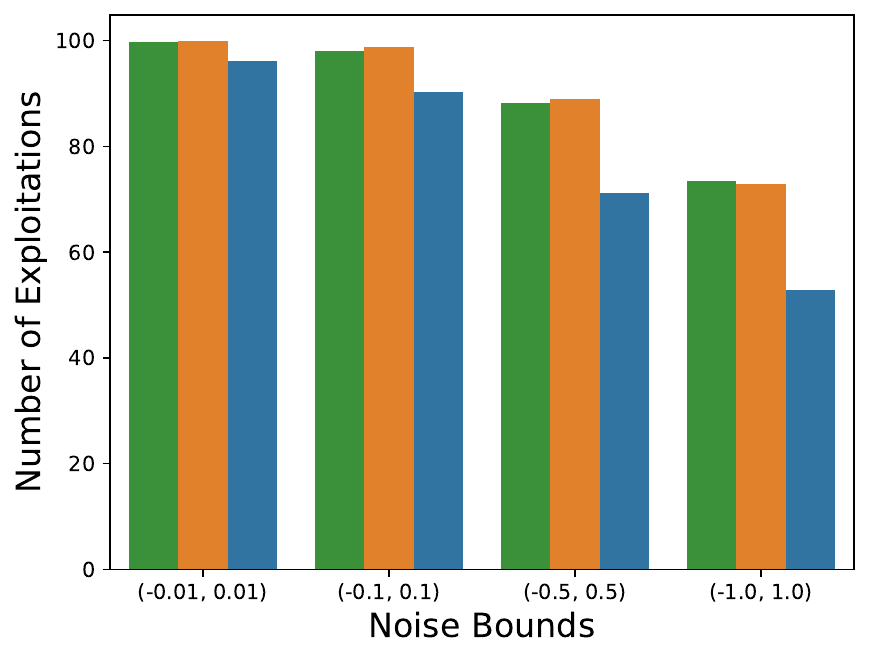}
    \caption{}
    \label{fig:db_attacks1}
    \end{subfigure}
    \begin{subfigure}[b]{0.32\textwidth}
    \includegraphics[width=\textwidth]{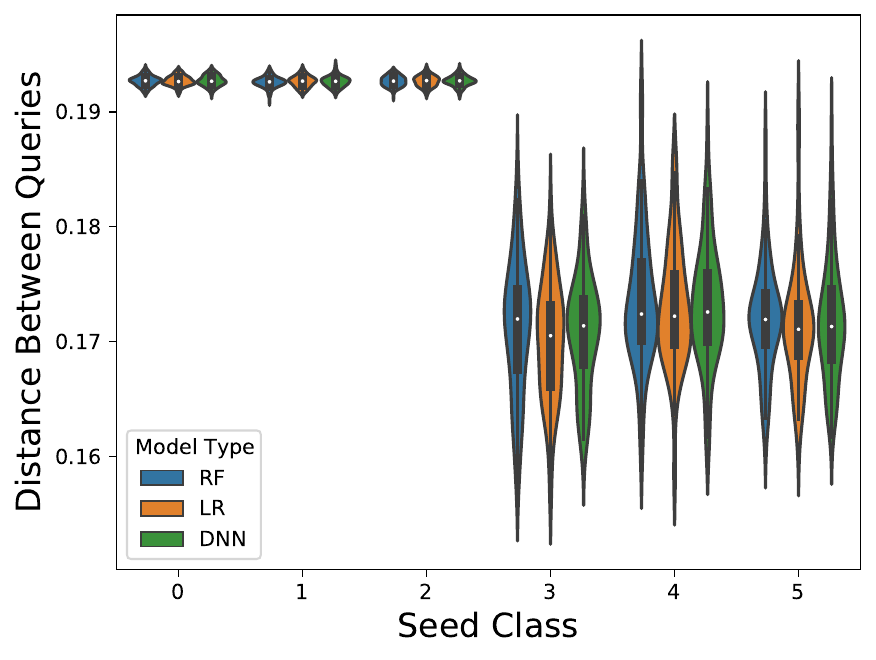}
    \caption{}
    \label{fig:db_attacks2}
    \end{subfigure}
    \begin{subfigure}[b]{0.32\textwidth}
    \includegraphics[width=\textwidth]{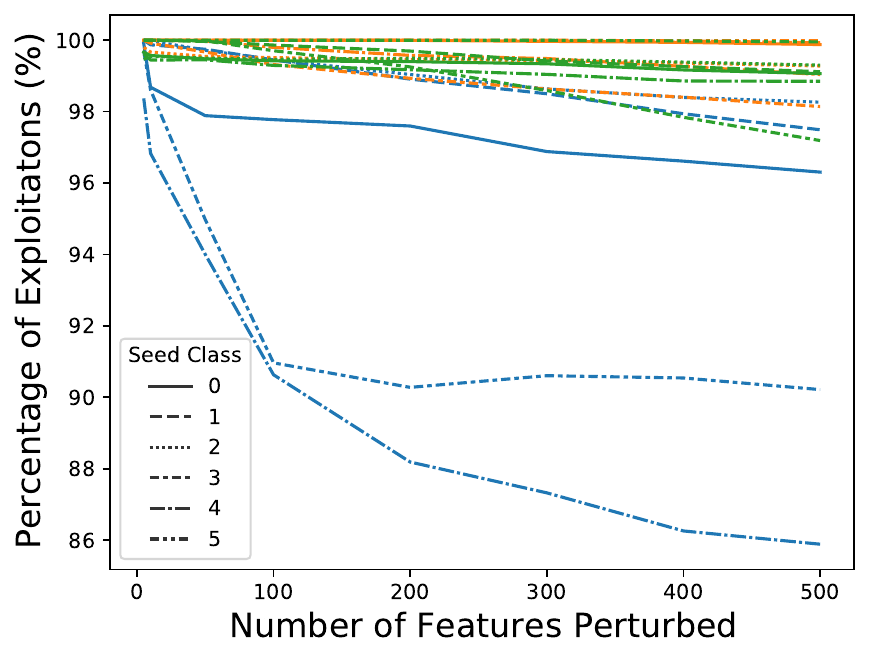}
    \caption{}
    \label{fig:db_attacks3}
    \end{subfigure}
    \caption{Results of random noise perturbation attacks to exploit decision boundaries of decision tree (DT), multi-class logistic regression model (LR) and deep neural network (DNN): (a) impact of noise bounds on the number of exploitations; (b) euclidean distance between exploitable queries for each seed class; and (c) impact of the number of features perturbed on the percentage of exploitations. The legend in (b) applies to all figures.}
    \label{fig:db_attacks}
\end{figure*}

\subsubsection{Exploiting Output Diversity (A-1)}

For adversarial goal A-1, we attempt to recover the prediction space (e.g., classes) beyond the seed query by drawing random queries from the uniform distribution. This attack can be executed by introducing randomness to either a subset or all of the features. This is equivalent to exploiting the input to receive maximally benefiting services such as highly profitable financial offers. Table \ref{tab:attack_results} contains the percentage of classes recovered via 100 random queries. While random forests are more robust to the attack with only 66.33\% and 29.50\% attack performance on HAR and MNIST respectively, the logistic regression and neural network models leak more than 90\% of the classes on both datasets.

We further examine the impact of varying attack parameters on the HAR dataset. Figure \ref{fig:class_attacks1} demonstrates the impact of query size on the percentage of classes recovered. Despite the simplicity of the attacks, we see 50\%-100\% leakage for different models across different number of query sizes. We also note the lower attack performance of random forests and attribute it to the robustness and unsmooth decision boundaries of ensemble modeling. Yet, we still see a $\sim 67$\% leakage with random forests for 5000 queries. 

Although there is limited distinction in leakage of white-box versus black-box models, there is a substantial difference in runtimes. Table \ref{tab:class_attack_runtime} contains the runtimes of recovering the maximum percentage of prediction space before plateauing. With only 1000 extra features, the runtime of the black-box attack was up to 199 times slower. If the percentage of features used for model training is substantially lower than the total features collected, the runtime increases substantially for black-box environments. This is shown in Figure \ref{fig:class_attacks3} where an addition of 10,000 extra features can take $\sim30$ hours to run for 5000 queries. In the white-box setting, the adversary has access to the model inputs. However, with large model input sizes, the attack may be difficult. 

Figure \ref{fig:class_attacks2} demonstrates the impact of randomly selecting a subset of the features on the prediction space recovered. The random selection of the subset of features is averaged across 10 samples for 1000 queries. Even with querying only half of the features, there is up to $\sim70\%$ leakage\footnote{This may differ for different datasets depending on how important certain features are for the model.}.

\vspace{0.5em}
\textbf{Key Takeaway: } Randomly generated queries can expose up to 100\% of model outputs, with random forests' robustness attributed to their unsmooth decision boundaries. However, black-box model attacks, while equally effective, have significantly longer runtimes.
%Randomly generated queries can successfully reveal up to 100\% of the model outputs with random forests being more robust due to their inherent unsmooth decision boundaries. While attacks on black-box and white-box ML models demonstrate similar performance, the runtime is substantially higher for black-box models.

\subsubsection{Exploiting Decision Boundaries (A-2)}

For adversarial goal A-2, we attempt to exploit the decision boundary of the seed query class. This is similar to identifying ways that the input can be perturbed while achieving the same service. Prior work in the model stealing literature has employed out-of-distribution (OOD) surrogate queries to exploit decision boundaries of the target model and train a clone model \cite{orekondy2019knockoff, papernot2017practical, juuti2019prada}. However, many of these sophisticated querying attacks only work for deep learning models with the end goal of cloning the target model. We focus on exploitation of decision boundaries by maximizing the number of different queries of the same class without the need to cover any particular space.

To distort the seed input, we add random noise drawn from a uniform distribution to a subset or all the features. Table \ref{tab:attack_results} contains results of adding noise drawn from [-0.01, 0.01] to all the features. The results are averaged across all classes. For both datasets, small perturbations lead to a high exploitation accuracy ranging from 96.18\% to 100.00\%. As in the previous attack, the LR and DNN models are slightly more vulnerable than the RF model.

We further examine the impact of attack parameters on the HAR dataset. Figure \ref{fig:db_attacks1} demonstrates the impact of noise bounds on the number of exploitations (i.e., same class queries) made. With smaller noise perturbations, exploitations are much easier to conduct as expected. Similar to the previous attack however, the random forest is more robust than the DNN and logistic regression model. With higher noise bounds, the random forest successfully exploits through only $\sim50$\% of the queries. 
% Figure \ref{fig:db_attacks1} demonstrates the impact of query size on the number of exploitations (i.e., same class queries) made. Similar to the previous attack, the DNN and logistic regression model perform worst than the random forest for all seed classes. For example, with 5000 queries for seed 5, the DNN and logistic regression model had 3687 and 3911 exploitations whereas the random forest only had 1834 exploitations. We also note the easabilitity with which certain classes are exploitable and that remains consistent across the models.

Figure \ref{fig:db_attacks2} considers the difference in exploited queries for different seed classes with noise generated from [-0.01, 0.01] bound. The required perturbations to exploit a particular class differ; while classes 0-2 require larger differences in perturbations, classes 3-5 allow more variance in perturbations leading to easability in the attack. Exploitations thus depend on the nature of the data and the impact of features on each class. 

As considered in the last attack, we also consider the impact of only perturbing a subset of features. Results in Figure \ref{fig:db_attacks3} demonstrate the lowering efficacy of the attack as more features are queried. Intuitively, this suggests that a smaller change yields the same result whereas perturbing large number of features can change the model output. This reasoning makes it much easier to conduct such an attack, both from an efficiency and latency standpoint.

\vspace{0.5em}
\textbf{Key Takeaway: } Perturbation attacks on decision boundaries achieve up to 100\% success with smaller bounds. While perturbing fewer features yields higher exploitation success, the impact on different classes varies.
%Exploiting decision boundaries via perturbation attacks can achieve up to 100\% success rate with smaller perturbation bounds. While perturbing a smaller number of features reveals higher exploitation success, the impact of perturbed queries on different classes can vary.

%More specifically, x did this, y did this and z did this... 
% While the former adversarial goal requires exploitation through OOD queries around the decision boundaries of the target model, goal A2 may be exploited with in-distribution (ID) queries as well.

% Consider the example in Figure \ref{} which shows a decision boundary of a binary classifier trained on an artificial dataset with two features. In order to exploit A1, an adversary has to use queries closer to the decision boundaries. On the contrary, A2 can suffice with either ID or OOD queries, making it even simpler.

% \textbf{Datasets.} Adobe Target (4 output classes -- DNN, RF), xxx (1 output class -- DNN, RF), and mnist (10 output classes -- DNN only)

% Goal: reconstruct function F(x) = y using F(x') = y queries where x' ~ x.

% Attacks themselves are not unique but demonstrating the leakage they can cause in on-device applications is unique.

% \textbf{Models.} One-vs-all (traditional) and one-vs-all neural networks where each output node represents a class (https://developers.google.com/machine-learning/crash-course/multi-class-neural-networks/one-vs-all). Use random forests and neural networks.

\section{Preserving Proprietary Information in On-Device Models}

Deploying models on the edge device of the end user reduces the service provider’s control of its usage and makes it difficult to track or identify adversarial actions. In this section, we propose an end-to-end modeling system, \system, for deploying and serving ML models on user devices while protecting the proprietary information embedded in the model through an in-built defense layer.

In design of \system, we account for several non-trivial challenges:
\begin{enumerate}
    \item \textbf{Nature of Adversarial Queries:} As demonstrated by our simple attacks in Section \ref{sec:attack}, the nature of adversarial queries can vary from simple single feature perturbations to complex randomly generated queries. This randomness can lead to both in-distribution (ID) queries and out-of-distribution (OOD) queries. While prior works have considered detecting OOD queries \cite{juuti2019prada, kariyappa2020defending, kariyappa2021protecting}, we make no assumptions on the distribution of the queries. 
    \item \textbf{Nature of Adversarial Goals: } As discussed in Section \ref{sec:threat-model}, adversarial goals can fall into one of two categories: (1) exploiting output diversity, and (2) exploiting decision boundaries. The nature of output expected for both these goals is opposite, requiring a complex solution that does not depend solely on output leakage for adversarial detection.
    \item \textbf{Nature of On-Device Models: } On-device serving often supports offline access and preserves user data privacy, a major deterrence to relying on the cloud for identifying adversarial actions. The system design must thus be able to run end-to-end on the device with requiring minimal to no cloud support.
\end{enumerate}

\begin{figure}
    \centering
    \includegraphics[width=\columnwidth]{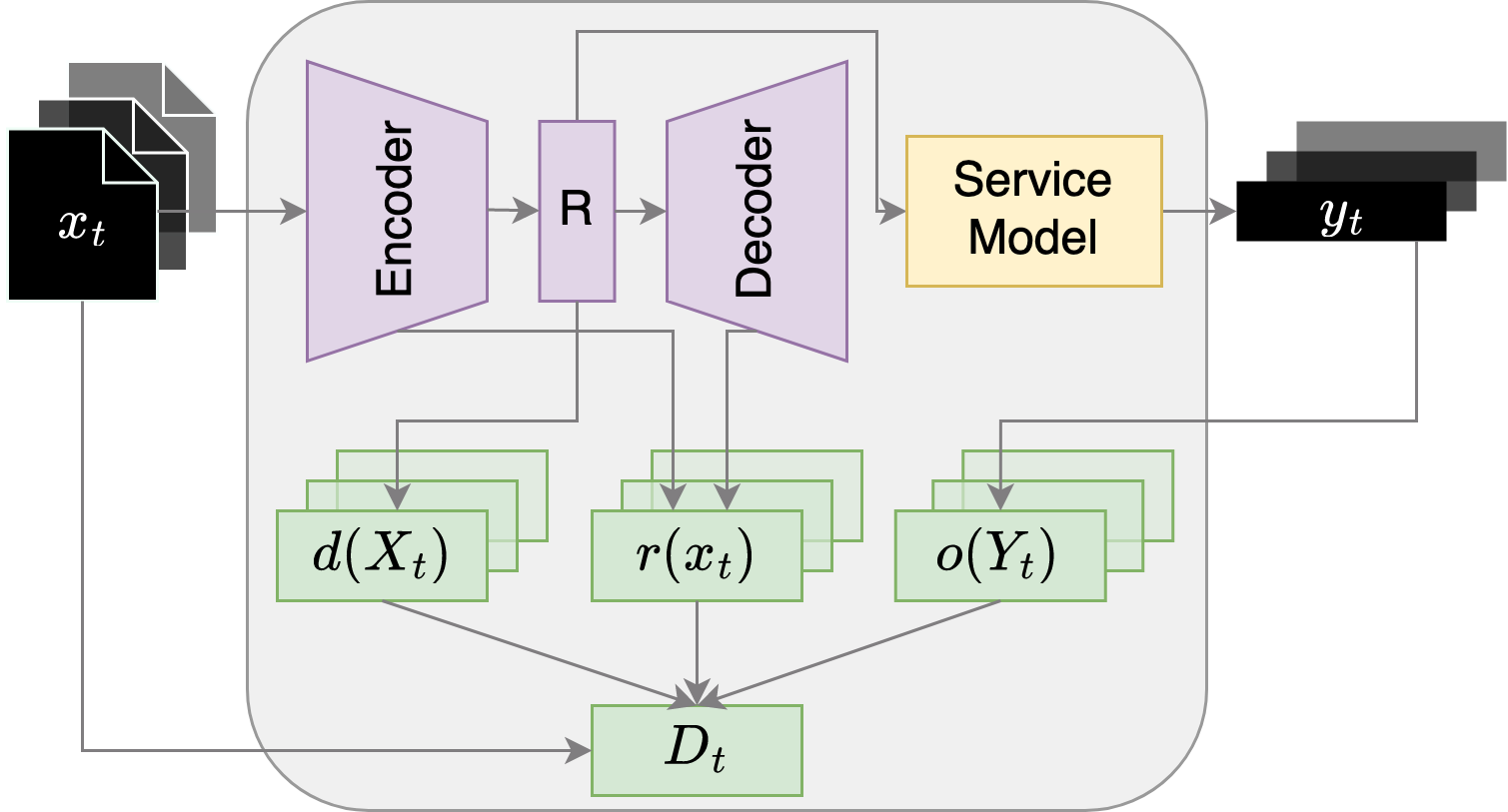}
    \caption{Overview of the proposed inference pipeline of \system. The purple blocks represent the autoencoder, yellow block represents the service model, and the green blocks represent the detector mechanism where detector output $D_t$ is an aggregation of the query distance, $d(X_t)$, reconstruction error, $r(x_t)$, and output entropy, $o(Y_t)$. Processing in the gray box occurs in memory while the application is running on the edge device and is assumed to be inaccessible by the user.}
    \label{fig:system_design}
    % \vspace{-2em}
\end{figure}

To satisfy the above challenges, \system's architecture is designed to maintain requirements of on-device models such as low latency, privacy of user data, and offline access, while defending against attacks that leak output diversity and decision boundaries of classes. The defense layer comprises of an autoencoder that is used to detect adversarial usage and label successive queries as benign or adversarial. Figure \ref{fig:system_design} demonstrates the design of \system. 

\system consists of the following key components. 

\subsection{Autoencoder Training}
\label{sec:autoencoder}
The first step is to train an autoencoder to convert the input query into a lower dimensional representation. The training can occur on the cloud or edge network, unknown from the user. The autoencoder consists of an encoder $f_e$, which takes as input a k-dimensional query $x_i$ and converts it into a m-dimensional vector $x_e$ where $m < k$, and a decoder $f_d$, which takes as input the output of the encoder and reconstructs the original k-dimensional query $x'_i$.
\begin{align*}
    f_e(x_i): \mathbb{R}_k \to \mathbb{R}_m \\
    f_d(x_e): \mathbb{R}_m \to \mathbb{R}_k
\end{align*}
The autoencoder is trained to minimize the mean squared error loss over the training data, $X_{\text{train}}$ as follows:
\[L(X_{\text{train}}) = \frac{1}{|X_{\text{train}}|}(X - f_d(f_e(X_{\text{train}})))^2\]

Autoencoders are a popular anomaly detection method as they fail to reconstruct anomalous inputs by design \cite{zhou2017anomaly}. This behavior stems from training the autoencoder on a dataset comprised mostly of benign instances, thereby enabling it to accurately reconstruct such instances. When confronted with anomalous inputs, which differ significantly from the patterns learned during training, the autoencoder struggles to faithfully reconstruct them, resulting in a higher reconstruction error. 
%During training, the autoencoder learns to extract key properties from the training data and reconstruct the original input from those key properties. 
The reconstruction error between the original input $X_i$ and reconstructed input $X'_i$ is expected to be low for ID queries and high for OOD queries. We use this aspect of autoencoders during the design of the defense layer (see Section \ref{sec:defense_layer}).

\subsection{Service Model Training}
\label{sec:service_model}
The second step involves training the model that is used to aid an arbitrary service (e.g., offer recommendation). As before, the training of the service model can occur on the cloud or edge network, unknown from the user. This model takes as input the output of the encoder $f_e(X_{\text{train}})$ and outputs a service $Y$. 
\[M(f_e(X_{\text{train}})): \mathbb{R}_m \to \mathbb{}R_c \]
The encoder maps the input data into a lower-dimensional representation, often referred to as the latent space, that captures essential features and patterns of the input data. Integrating the training of model $M$ with an autoencoder enables $M$ to leverage the learned latent representation and fine-tune it specifically for the targeted service, eliminating the need for relearning from raw input data. By design, the model can be of any complexity, ranging from simple linear models to more complex deep networks. 
    
\subsection{Adversarial Detection}
\label{sec:defense_layer}
The third step of the framework is adding the defense mechanism that guards against adversarial usage of the models described above. \system's defense mechanism is built upon the measurement of leakage rate ($l$) across time.
% \begin{align}
%     D_t = \frac{l_t}{s_t}
% \end{align}
% Particularly, at each time $t$, we measure the amount of leakage that has occurred and compare that with how often the service has been queried. If both the leakage and service rate are high, this suggests that the service is being used in an adversarial manner. On the other hand, if leakage rate is low, this suggests that the service is being used by a benign user. Finally, if the leakage rate is high but the service rate is low, this can raise a suspicious flag. The thresholds that allow differentiation between adversarial, suspicious and benign users will be application-dependent.

In order to capture the challenges with building a generalized and robust solution towards protecting proprietary information, we define leakage rate using three components. First, we capture OOD queries via the autoencoder's reconstruction error $r(x_t)$ where $x_t$ is the k-dimensional query at time $t$. As discussed in Section \ref{sec:autoencoder}, we exploit the fact that autoencoders are unable to reconstruct queries coming from a different distribution than the training data. We use cumulative mean squared error to calculate a moving total of the error between the original query and the reconstructed query (e.g., output of the decoder). 
\begin{align}
    r(x_t) = r(x_{t-1}) + \frac{1}{k}\sum_i^k \left( (x_t^i - f_d(f_e(x_t))^i)^2 \right)
\end{align}

The second component in the computation of leakage rate is distance between the queries $d(X_t)$ where $X_t$ represents the set of queries $\{x_1, x_2,...,x_t\}$ until time $t$. Intuitively, we expect queries from adversaries exploiting decision boundaries to be very similar with minute perturbations whereas adversaries exploiting output diversity will have fairly different queries. By measuring the distance between the queries, we capture queries that are too similar or too different. We use the cumulative median Euclidean distance between encoded query at time $t$ and all the previous encoded queries by the same user to identify the distance at time $t$. Euclidean distance is chosen as a measure of distance due to its ability to measure magnitude well.
% \begin{align}
%     d(X_t) = d(X_{t-1}) + \frac{1}{t}\sum_i^{t-1} \lVert f_e(x_i) - f_e(x_t) \rVert
% \end{align}
\begin{align}
    d(X_t) = d(X_{t-1}) + \notag \\
    \text{med}(\lVert f_e(x_i) - f_e(x_t) \rVert \text{ } \forall i \in [0,t\!-\!1] )
\end{align}

The third component in the computation of leakage rate is the output entropy $o(Y_t)$ where $Y_t$ are the set of predictions of the service model $\{y_1, y_2,...,y_t\}$ until time $t$. Entropy is a measure of the randomness or uncertainty. We use this as a measure to explore the diversity of the predictions made by the model. Intuitively, substantially high entropy would indicate diverse and varied predictions, whereas substantially low entropy would indicate similar or repetitive predictions, both of which can be suggestive towards adversarial usage. We can use the extremeness of the entropy as an indicator of adversarial usage. Let $p_i$ represent the probability of occurrence of class $c_i$. Then, we have:
\begin{align}
    o(Y_t) = -\sum_i^C p_i\log(p_i)
\end{align}

All components undergo min-max normalization with respect to the training data. This normalization technique scales each component's values with respect to the minimum and maximum values observed in the training data at each time step. By applying this normalization process, the components are transformed to a standardized scale that facilitates fair and meaningful comparisons. The final computation of leakage rate $l_t$ is an aggregation of the three components above as follows:
\begin{align}
    \label{eq:leakage_rate}
    l_t = \alpha r(x_t) + \beta d(X_t) + \gamma o(Yt)
\end{align}
where $\alpha$, $\beta$ and $\gamma$ are the weights associated with the cumulative reconstruction error, cumulative median euclidean distance and output entropy respectively such that $\alpha + \beta + \gamma = 1$.

Finally, the detector categorizes queries observed until time $t$ as either benign or adversarial through threshold scaling as follows:
\begin{align}
    D_t = 
\begin{cases}
    \label{eq:detection}
    1 , & \text{if } l_t < l^{tr}_t - \delta l^{tr}_t \text{ or } l_t > l^{tr}_t + \delta l^{tr}_t \\
    0,  & \text{otherwise}
\end{cases}
\end{align}

Here, $l_t^{tr}$ represents the leakage rate of the training data at time $t$. If the leakage rate $l_t$ falls outside the range defined by $l^{tr}_t - \delta l^{tr}_t$ and $l^{tr}_t + \delta l^{tr}_t$, the query is classified as adversarial (1). Otherwise, it is classified as benign (0). The selection of an appropriate value for $\delta$ depends on the specific requirements and characteristics of the data. Upon detection of adversarial usage against the adversary, the system delegates the responsibility of choosing the appropriate action, such as blocking further use or implementing periodic suspension, to the service provider.

\subsection{System Deployment}
\begin{figure}
    \centering
    \includegraphics[width=\columnwidth]{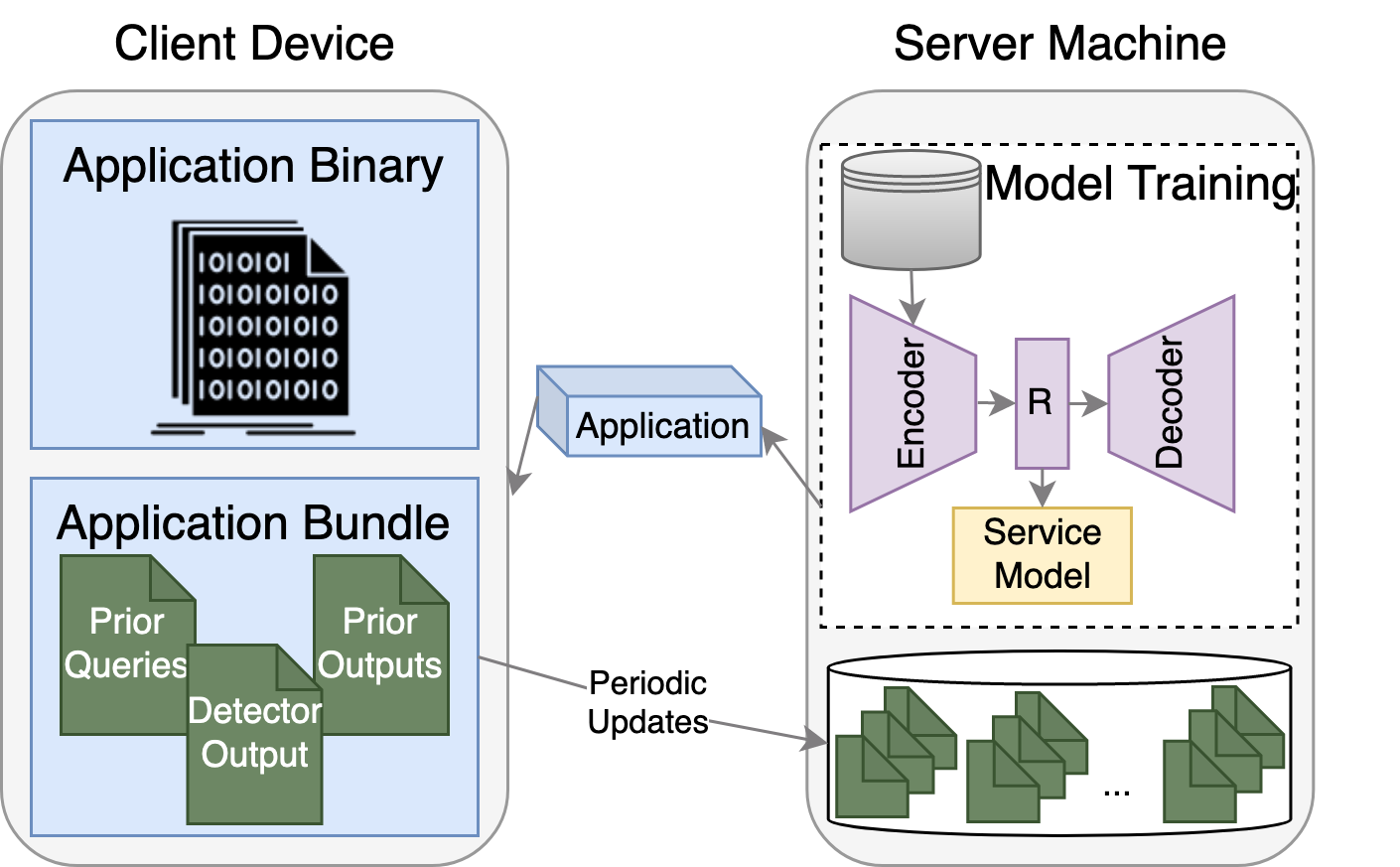}
    \caption{Overview of the model training and deployment in \system. The application binary contains the program itself which runs in memory when the application is invoked. The application bundle contains the files the application uses, including list of encoded queries, prior model outputs and prior values of the adversarial detection layer.}
    \label{fig:model_deployment}
\end{figure}

The fourth and final step of \system enables secure system deployment on the device where the models are deployed in a black-box manner. Figure \ref{fig:model_deployment} demonstrates the design of the training and deployment process. The training steps from Sections \ref{sec:autoencoder} and \ref{sec:service_model} occur on the server machine (e.g., cloud). Once trained, \system deploys the model on the user's device in a black-box manner by either embedding the model into the binary of the application or by storing the model as encrypted serialized model files. For the latter method, we use the Open Neural Network Exchange (ONNX) format to serialize model files as it is interoperable and framework-independent \cite{onnx2022}. 

The model parameters (e.g., weights) and other relevant files from Section \ref{sec:defense_layer}, including a list of encoded queries, prior model outputs, and prior values of the adversarial detection layer, are encrypted and stored on the device. We employ the XSalsa20 stream cipher for encryption of these files. XSalsa20 provides a high level of security and efficiency while being well-suited for devices with limited computational power. The encryption and decryption happens in memory while the application is running and is expected to be secure from adversarial access. 
%To further enable protection of these files on the device, we use access control lists to restrict users from accessing or modifying these files.

Finally, \system periodically uploads these data files to the cloud to ensure recovery during deletion or application reset.\footnote{Under constraints of complete offline access, enabling user authentication and file monitoring can achieve the same goal.} This ensures that adversaries cannot delete these files or the application and restart their attack from scratch. The frequency of the uploads along with the nature of the file monitoring can be determined by the service provider. Since the data being sent to the cloud does not contain raw queries made by the user, user data is still considered to be preserved.

\section{Prototype Implementation}

We implement a prototype of the proposed system \system on a Raspberry Pi 3 (RPI) running Debian GNU/Linux 11. The RPI is equipped with a 64-bit quad-core ARM Cortex-A53 processor and 1GB RAM. The prototype is developed using Python 3.6. The remainder of this section describes our implementation of \system including model training, adversarial detection, and optimizations for on-device deployment.

\vspace{0.5em}
\textbf{Model Training.} \system comprises of two types of models, autoencoder and service model, as described in Sections \ref{sec:autoencoder} and \ref{sec:service_model} respectively. The autoencoder is instantiated within the \emph{PyTorch} framework. It consists of two \texttt{nn.Sequential()} modules, designed to serve as the encoder and decoder components. Each of these modules comprises two linear layers with a ReLU activation. Additionally, the prototype accommodates three variants of service models: random forests (RF), multi-class logistic regression (LR), and deep neural network (DNN). The DNN model, implemented using PyTorch, is structured with three linear layers interspersed with ReLU activations. Conversely, the RF and LR models are implemented using the \emph{Scikit-Learn library}. The selection of optimal hyperparameters for each model is achieved through randomized search employing 3-fold cross-validation. The model training process is performed on an NVIDIA Titan-X GPU equipped with 32GB memory.

\vspace{0.5em}
\textbf{Adversarial Detection.} The adversarial detection, as described in Section \ref{sec:defense_layer}, involves categorizing the leakage rate via query distance, autoencoder reconstruction error, and output entropy. We leverage the \emph{threading} library to implement multithreading, allowing for concurrent computation of the leakage rate components. Additionally, to safeguard against tampering with the output files generated during this process, we employ the \emph{PyNaCl} library, which enables encryption through the XSalsa20 stream cipher and authentication via the Poly1305 MAC mechanism.

\begin{table}[t!]
    \centering
    \small
    \caption{Difference between traditional serialized models and ONNX models for single query inference (aggregated across 100 queries). AE refers to the autoencoder model used in SODA.}
    \begin{tabular}{c|c|c|c|c}
         & \textbf{Serialization} & \textbf{Performance} & \textbf{Runtime (ms)} & \textbf{Size (KB)} \\
         \hline
         \multirow{2}{*}{\textbf{AE}} & Traditional & 5.75e-3 & 0.237 & 2834 \\
         & ONNX & 5.86e-3 & 0.154 & 726 \\
         \hline
         \multirow{2}{*}{\textbf{RF}} & Traditional & 88.00\% & 13.749 & 4474 \\
         & ONNX & 88.00\% & 0.250 & 2903 \\
         \hline
         \multirow{2}{*}{\textbf{LR}} & Traditional & 92.00\% & 0.217 & 7 \\
         & ONNX & 92.00\% & 0.058 & 5 \\
         \hline
         \multirow{2}{*}{\textbf{DNN}} & Traditional & 92.00\% & 0.268 & 138 \\
         & ONNX & 92.00\% & 0.087 & 37 \\
    \end{tabular}
    \label{tab:onnx-performance}
\end{table}

% -------------AE-------------
% ONNX Performance:  0.005858057611308496
% ONNX Inference time:  0.0001544356346130371
% ONNX Storage: 726KB
% PyTorch Performance:  0.005753184653614314
% PyTorch Inference time:  0.00023668527603149414
% PyTorch Storage: 2.8MB
% -------------RF-------------
% ONNX Performance:  0.88
% ONNX Inference time:  0.000249636173248291
% ONNX Storage: 2.9MB
% Sklearn Performance:  0.88
% Sklearn Inference time:  0.013749396800994873
% Sklearn Storage: 4.5MB
% -------------LR-------------
% ONNX Performance:  0.92
% ONNX Inference time:  5.789041519165039e-05
% ONNX Storage: 5KB
% Sklearn Performance:  0.92
% Sklearn Inference time:  0.00021722793579101563
% Sklearn Storage: 7KB
% -------------DNN-------------
% ONNX Performance:  0.92
% ONNX Inference time:  8.676767349243164e-05
% ONNX Storage: 37KB
% PyTorch Performance:  0.92
% PyTorch Inference time:  0.00026826858520507814
% PyTorch Storage: 138KB

\vspace{0.5em}
\textbf{System Deployment.} The deployment of the application inference pipeline adheres to the architectural design depicted in Figure \ref{fig:system_design}. To achieve interoperability, we employ the \emph{skl2onnx} library and the \texttt{torch.onnx.export()} function to serialize the models into the ONNX format. Furthermore, in order to further minimize the memory footprint of PyTorch models, dynamic quantization is employed. This process converts single precision model parameters into reduced precision integer representation, without significant loss in accuracy. The efficacy of utilizing ONNX runtime in comparison to the Scikit-Learn and PyTorch frameworks is demonstrated in Table \ref{tab:onnx-performance}, showcasing substantial improvements in runtime and model size. Lastly, the inference pipeline is packaged into an executable file via the \emph{Pyinstaller} library.

By design, when the application prototype is invoked, the detector processes the input and updates resource files in the application bundle. If the detector detects adversarial usage, the user is temporarily suspended for a designated time period. To ensure data integrity, file updates are sent to another server at frequent intervals. Finally, file monitoring is enabled such that if files in the application bundle get deleted, the application won't execute the request and the user will be blocked for a time period.

% Steps for constructing prototype:
% \begin{enumerate}
%     \item Train models on cloud.
%     \item Convert models to ONNX format for interoperatability and optimize for low resource devices (Table \ref{tab:onnx-performance}).
%     \item Write deployment script which takes as input a query and outputs a result (e.g., service). 
%     \item Convert deployment script into a bundle that consists of an executable, dependencies, and resource files.
%     \item Deploy the application on a Raspberry Pi.
%     \item Invoke application by running the executable with command-line arguments (e.g., query).
% \end{enumerate}

% Prototype features:
% \begin{enumerate}
%     \item When the application is invoked, the detector processes the input in the background and updates resource files in bundle. File updates are sent to the cloud at frequent intervals (e.g., every 10 queries).
%     \item The resource files are saved with encryption.
%     \item File monitoring is enabled such that if a file gets deleted, the application won't execute the command and the user will be blocked for a time period.
%     \item If the detector outputs adversarial, the user is blocked for a time period.
%     \item Models are updated frequently (e.g., daily or weekly).
% \end{enumerate}

\begin{table*}[t!]
    \centering
    \small
    \caption{Performance of adversarial detection in the proposed system as compared to random baseline and prior works.}
    \begin{tabular}{c|l|ccc|ccc|ccc}
        \hline
         \textbf{Dataset} & \textbf{Method} & \multicolumn{3}{c}{\textbf{Random Forest}} & \multicolumn{3}{c}{\textbf{Logistic Regression}} & \multicolumn{3}{c}{\textbf{Deep Neural Network}} \\
         & & \textbf{Accuracy} & \textbf{Precision} & \textbf{Recall} & \textbf{Accuracy} & \textbf{Precision} & \textbf{Recall} & \textbf{Accuracy} & \textbf{Precision} & \textbf{Recall} \\
         \hline
         \multirow{4}{*}{HAR} & Random & 57.69 & 81.69 & 58.00 & 46.92 & 73.85 & 48.00 & 52.31 & 78.79 & 52.00 \\ 
         & MagNet* & 75.00 & 100.00 & 50.00 & 75.00 & 100.00 & 50.00 & 75.00 & 100.00 & 50.00 \\
         & PRADA* & 77.69 & 77.52 & 100.00 & 77.69 & 77.52 & 100.00 & 78.46 & 78.13 & 100.00 \\
         % & Output Entropy & 56.154 & 87.719 & 50.000 & 76.154 & 89.655 & 78.000 & 93.077 & 92.523 & 99.000 \\ 
         % & PRADA & \\
         & \textbf{\system} & 84.62 & 90.00 & 90.00 & 89.23 & 91.35 & 95.00 & 88.46 & 91.26 & 94.00 \\
         \hline
         \multirow{4}{*}{MNIST} & Random & 49.09 & 33.87 & 42.00 & 49.82 & 36.62 & 52.00 & 49.46 & 34.15 & 42.00 \\ 
         & MagNet* & 75.00 & 100.00 & 50.00 & 75.00 & 100.00 & 50.00 & 75.00 & 100.00 & 50.00 \\ 
         & PRADA* & 36.36 & 36.36 & 100.00 & 36.36 & 36.36 & 100.00 & 36.36 & 36.36 & 100.00 \\
         % & Output Entropy & 100.000 & 100.000 & 100.000 & 100.000 & 100.000 & 100.000 & 100.000 & 100.000 & 100.000\\ 
         % & PRADA* & \\
         & \textbf{\system} & 94.91 & 100.00 & 86.00 & 92.00 & 100.00 & 78.00 & 93.45 & 100.00 & 82.00 \\
         \hline
    \end{tabular}
    \label{tab:comparison_prior}
\end{table*}

\begin{figure*}[t!]
    \centering
    \begin{subfigure}[b]{0.32\textwidth}
    \includegraphics[width=\textwidth]{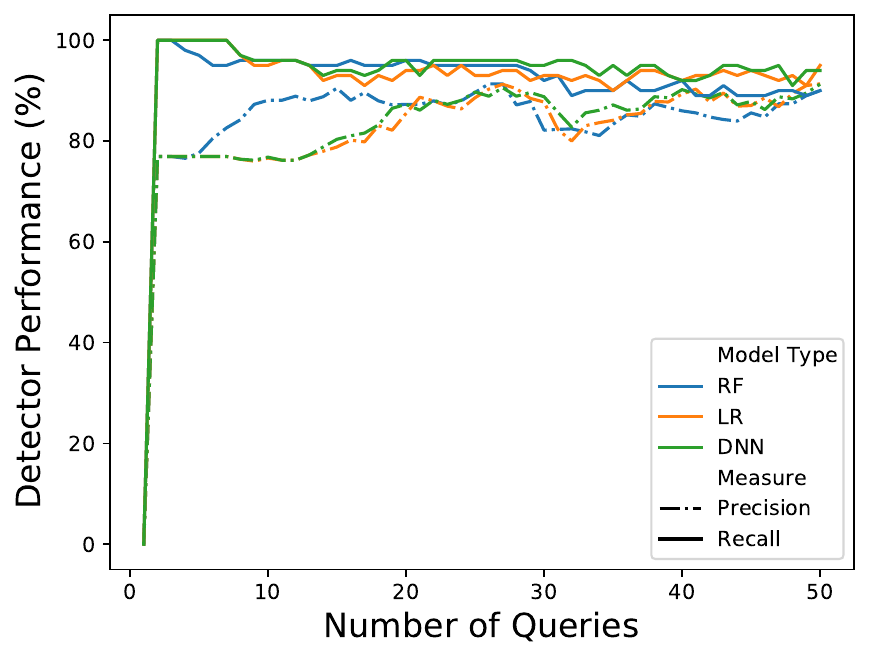}
    \caption{}
    \label{fig:performance}
    \end{subfigure}
    \begin{subfigure}[b]{0.32\textwidth}
    \includegraphics[width=\textwidth]{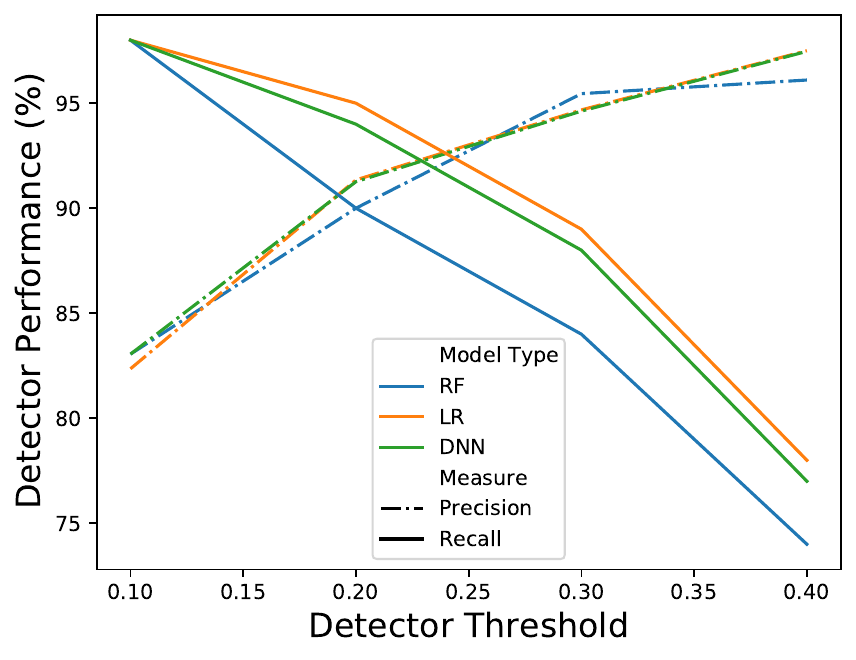}
    \caption{}
    \label{fig:thresholds}
    \end{subfigure}
    \begin{subfigure}[b]{0.32\textwidth}
    \includegraphics[width=\textwidth]{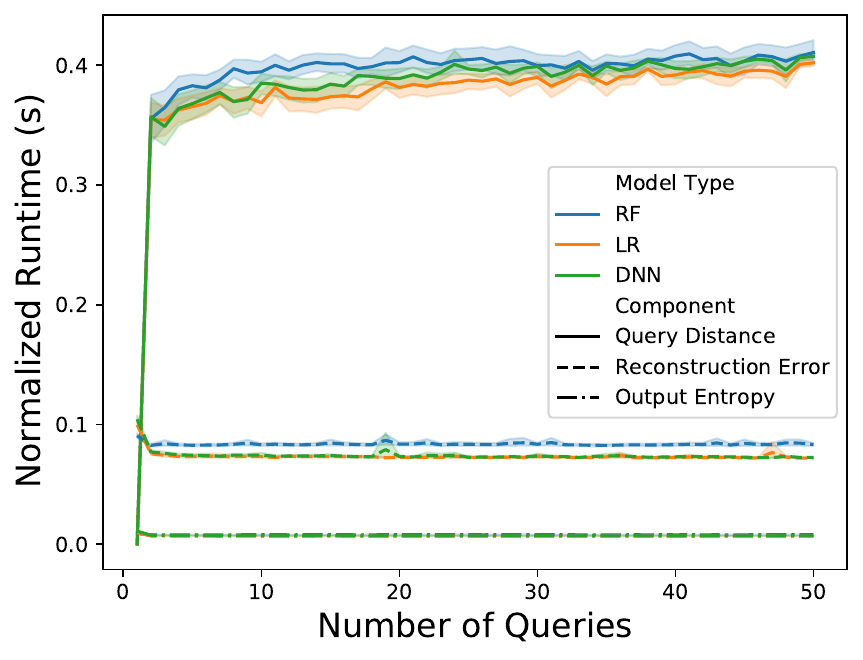}
    \caption{}
    \label{fig:runtime}
    \end{subfigure}
    \caption{Results of evaluating \system's output: (a) detector performance with respect to the number of queries (b) detector performance on 50 queries with respect to the detector threshold; and (c) runtime of the detector components with respect to the number of queries.}
    \label{fig:defense_performance}
\end{figure*}

\section{System Evaluation}
In this section, we demonstrate the efficacy of employing \system for on-device deployment on high-level objectives including the ability to defend against adversaries, latency, size, and performance. We employ the HAR and MNIST datasets described in Section \ref{sec:attack-setup} for evaluation. As before, all results are aggregated across 100 adversaries. Unless otherwise stated, the system parameters are set to $\alpha=0.33$, $\beta=0.33$, $\gamma=0.33$, and $\delta=0.2$.

\subsection{Efficacy of Adversarial Detection}
To evaluate the efficacy of the adversarial detection layer, we employ the training set and test set as representative examples of benign queries. For the generation of adversarial data, we follow the methodology outlined in Section \ref{sec:attack}, which involves the selection of 100 random seed queries and subsequent generation of random and perturbed queries.

\subsubsection{Evaluation of Detection Output}
Our evaluation of the detection layer compares \system with three other methods. We first compare with a random baseline which randomly labels users as benign or adversarial. The remaining two methods are chosen based on proximity to our work. Specifically, MagNet proposed a two-pronged defense against adversarial examples using an autoencoder-based approach \cite{meng2017magnet}. We employ MagNet's autoencoder with an error threshold of 0.01 and 0.04 for the HAR and MNIST datasets, respectively. These thresholds were chosen based on the highest value observed in the benign set, ensuring a false positive rate of 0. We also compare with PRADA which labels successive queries as adversarial based on the distribution of distances between the queries \cite{juuti2019prada}. We set the detection threshold at 0.95, signifying a greater level of confidence in the normalcy of the observed distribution. Note, these comparisons are based on re-implementations of both systems.

The detection results are compared in the form of accuracy, precision, and recall. A high precision value indicates that when the model predicts an instance as adversarial, it is likely to be correct. It is useful when the cost of false positives (misclassifying benign users as adversarial) is high. Alternatively, a high recall value indicates that the model is effective at identifying adversaries and minimizing false negatives. It is particularly important when the cost of false negatives (failing to detect adversaries) is high, as it ensures that a larger proportion of adversaries are correctly identified.

As tabulated in Table \ref{tab:comparison_prior}, \system outperforms the other methods for 50 queries. Specifically, by combining multiple components for detection, \system achieved $\approx36\%$, $\approx13\%$, and $\approx10\%$ improvements in accuracy over the baseline, MagNet, and PRADA methods, respectively, for the HAR DNN. While precision and recall are higher for the other methods, \system performs the most optimally on all three metrics. For instance, while MagNet has 100\% precision, it fails to recognize half the adversaries as indicated by the 50\% recall. Conversely, while PRADA has a 100\% recall, it misclassifies many benign users as adversarial as indicated by the 78.13\% precision. This can be attributed to the different design requirements of the methods. While PRADA emphasizes long-term detection for model extraction attacks, MagNet emphasizes detection of perturbation-based adversarial attacks. SODA is a more generalizable method which incorporates different components for detecting a wider range of attacks. Similar results are noted for the random forest and logistic regression models and MNIST dataset.

\textbf{Key Takeaway: } The proposed system outperforms prior methods that pursued similar goals by 10\%-59\%. The results solidify \system's ability to detect a wider range of attacks as shown in Section \ref{sec:attack}, thereby providing strong evidence about its generalizability.

\begin{figure*}[t!]
    \centering
    \begin{subfigure}[b]{0.32\textwidth}
    \includegraphics[width=\textwidth]{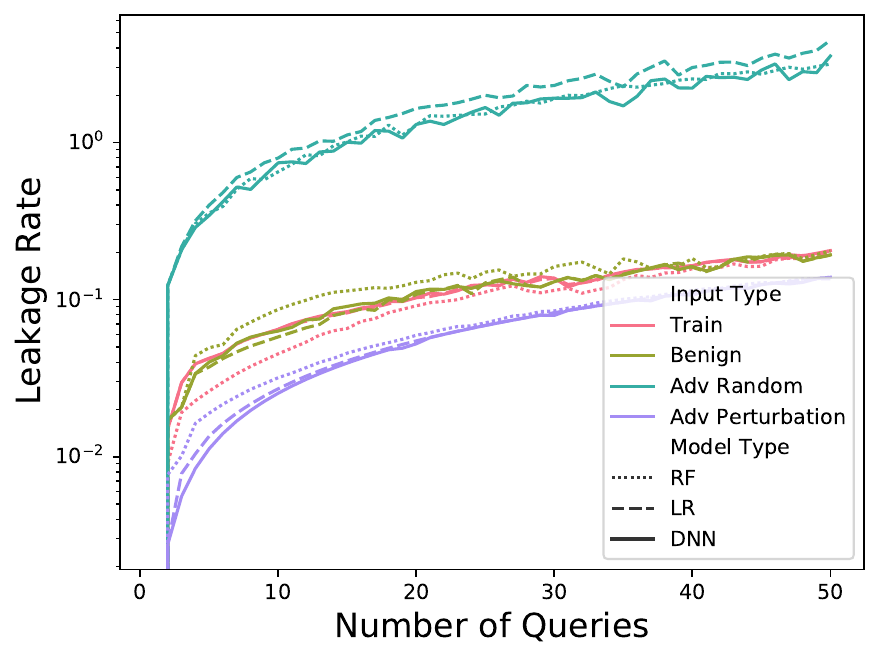}
    \caption{}
    \label{fig:leakage}
    \end{subfigure}
    \begin{subfigure}[b]{0.32\textwidth}
    \includegraphics[width=\textwidth]{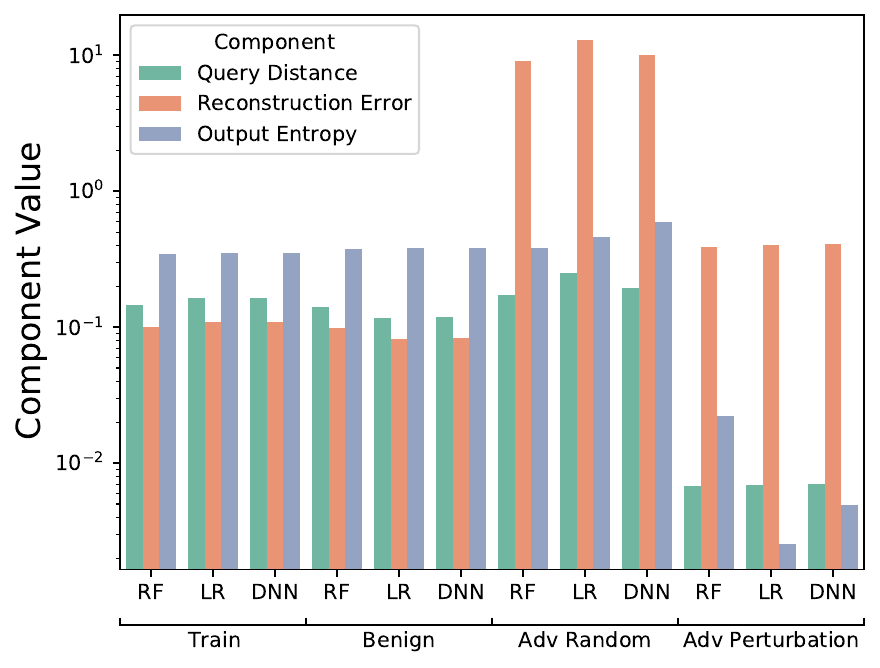}
    \caption{}
    \label{fig:leakage_component}
    \end{subfigure}
    \begin{subfigure}[b]{0.32\textwidth}
    \includegraphics[width=\textwidth]{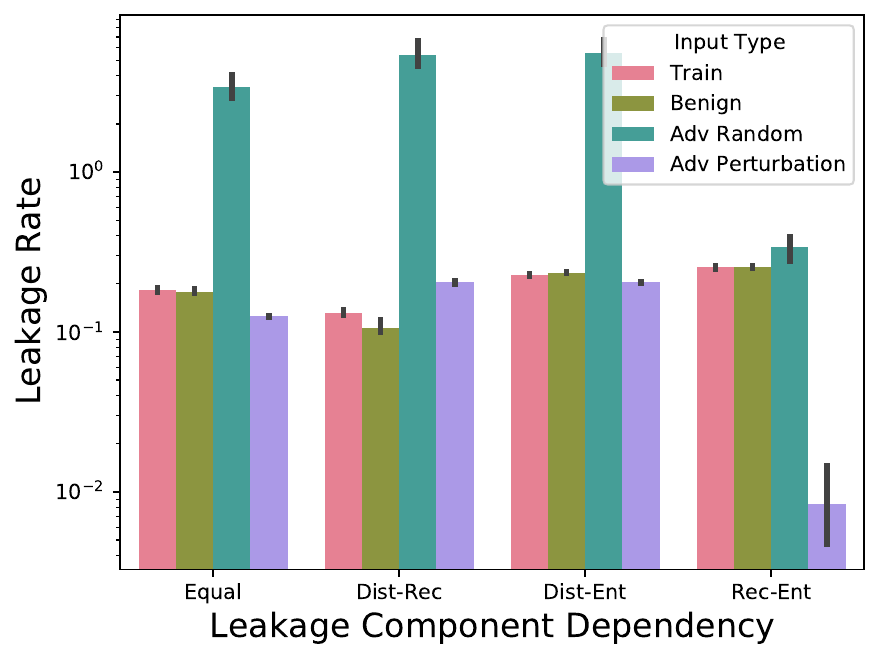}
    \caption{}
    \label{fig:leakage_thresholds}
    \end{subfigure}
    \caption{Results of analyzing leakage rate across input and model types: (a) leakage rate with respect to the number of queries across input and model types; (b) leakage component values across input and models types; and (c) affect of leakage component thresholds on leakage rate aggregated across models where `Dist` represents distance, `Rec` represents reconstruction error, and `Ent` represents output entropy.}
    \label{fig:defense_sanity}
\end{figure*}

\subsubsection{Impact of Detection Parameters}
We further evaluate resiliency of the defense mechanism across system parameters on the HAR dataset. Figure \ref{fig:performance} demonstrates the precision and recall values of the adversarial detection layer across queries. We observe that \system's ability to identify adversaries increases as the number of queries increases. It is important to note that precision and recall are typically inversely related. Increasing one often leads to a decrease in the other. Therefore, the choice between the two depends on the specific requirements of the system and the relative costs associated with false positives and false negatives. The effect described can be observed in Figure \ref{fig:thresholds}, which showcases the influence of the detector threshold $\delta$ (as defined in Equation \ref{eq:detection}). As $\delta$ increases, the recall decreases, resulting in an increase in false negatives. In our use case, a threshold value of $\delta=0.2$ strikes a balance between false positives and false negatives.

Figure \ref{fig:runtime} demonstrates the breakdown in system runtime across leakage rate components and number of queries. The computation of output entropy and reconstruction error exhibits consistent runtime, regardless of the number of queries. However, there is an increasing runtime observed for query distance computation due to the pairwise comparisons performed at each timestep. Additionally, we observe that the RF model has a higher runtime compared to the DNN and LR models, aligning with the overall longer runtime of RF inference demonstrated in Table \ref{tab:system_performance}.

\textbf{Key Takeaway: } Since \system is designed around the premise of comparing consequent queries, its effectiveness in detecting adversaries grows as more queries are provided, with minimal and consistent overhead in runtime across detection components. The ability to choose the detector threshold enables further flexibility in practical deployment.

\begin{table}[t!]
    \small
    \centering
    \caption{Inference overhead of proposed system \system compared to the traditional approach of on-device deployment.}
    %averaged across 10 queries
    \begin{tabular}{c|c|cccc}
         \textbf{Model} & \textbf{Method} & \textbf{Runtime (s) } & \textbf{Size (MB)} & \textbf{Accuracy (\%) }  \\
         \hline
         \multirow{2}{*}{RF} & Baseline & 2.086 & 40.51 & 92.50 \\
         & \system & 1.749 & 61.46 & 90.33 \\
         \hline
         \multirow{2}{*}{LR} & Baseline & 0.094 & 36.83 & 96.13 \\
         & \system & 0.234 & 58.56 & 92.94 \\
         \hline
         \multirow{2}{*}{DNN} & Baseline & 0.137 & 36.83 & 94.74 \\
         & \system & 0.291 & 58.59 & 92.90 \\
    \end{tabular}
    \label{tab:system_performance}
\end{table}

\begin{table*}[t!]
    \centering
    \small
    \caption{Summary of existing defenses against model extraction attacks comparing edge friendliness, model interoperability, and adaptability across adversaries.}
    \begin{tabular}{c|lccc}
        \textbf{Type} & \textbf{Method} & \textbf{Edge-Friendly} & \textbf{Interoperable} & \textbf{Adaptable} \\
        \hline 
        \multirow{4}{*}{Protection} & Protecting parameters with differential privacy \cite{tramer2016stealing} & \xmark & \cmark & \cmark \\
         & Training with adversarial diversity \cite{papernot2017practical, kariyappa2021protecting} & \xmark & \cmark & \xmark \\
         & Modifying prediction probabilities \cite{tramer2016stealing, lee2019defending, orekondy2019prediction} & \cmark & \cmark & \xmark \\
         & Adaptively misinforming for OOD queries \cite{kariyappa2020defending} & \cmark & \xmark & \xmark \\
         \hline
         
        \multirow{4}{*}{Detection} & Quantifying feature space explored 
        \cite{kesarwani2018model} & \cmark & \xmark & \cmark \\
         & Evaluating closeness to decision boundaries \cite{quiring2018forgotten} & \cmark & \cmark & \xmark  \\
         & Analyzing query distribution \cite{grosse2017statistical, meng2017magnet, juuti2019prada} & \cmark & \cmark & \xmark \\ 
         & \textbf{Analyzing queries and output leakage (\system)} & \cmark & \cmark & \cmark \\
         \hline
    \end{tabular}
    \label{tab:extraction-solutions}
\end{table*}

\subsubsection{Analysis of Leakage Rate}
To better understand the impact of leakage rate on the detection output, we perform an analysis of leakage rate across input and model types. Figure \ref{fig:leakage} showcases the impact of the number of queries on the leakage rate. While the leakage rate remains relatively consistent between the data obtained from model training and benign users across model types, a notable divergence occurs for both random and perturbed adversarial queries as the number of queries increases. This distinction corroborates the increasing recall seen in Figure \ref{fig:performance}.

A further breakdown of the component values in the leakage rate is demonstrated in Figure \ref{fig:leakage_component}. As described in Section \ref{sec:defense_layer}, query distance is important for identifying similarity or dissimilarity of queries over time, reconstruction error is important for identifying OOD queries, and output entropy is important for output diversity attacks. For random queries, the reconstruction error is significantly higher as compared to the error for training and benign data. Conversely, the query distance and output entropy are much lower for perturbed queries. This analysis suggests that \system effectively handles the differing nature of adversarial queries by leveraging different components in the leakage rate.

Lastly, we examine the influence of parameters in Equation \ref{eq:leakage_rate}. Visible differences are observed for random and perturbed queries when equal dependency is assigned to the parameters (e.g., $\alpha=0.33$, $\beta=0.33$, $\gamma=0.33$). Conversely, focusing on reconstruction error and output entropy only (e.g., $\alpha=0.5$, $\beta=0$, $\gamma=0.5$) suggests a larger distinction for perturbed queries and a smaller distinction for random queries. While the choice of parameter dependency can be tailored based on the system requirements, it is important to note that all three components of the leakage rate contribute value to the adversarial detection process.

\textbf{Key Takeaway: } The analysis of leakage rate reinforces the effectiveness of \system by revealing distinct detection patterns across benign and adversarial usage. While each component of the leakage rate contributes value to the detection of adversaries, it is possible to assign dependency to these components based on the application requirements.

%% full inference + detector results
% \begin{table}[]
%     \small
%     \centering
%     \caption{Overhead of proposed system \system compared to traditional approach of on-device deployment.}
%     %averaged across 10 queries
%     \begin{tabular}{c|c|ccc}
%          \textbf{Model} & \textbf{Method} & \textbf{Runtime (s) } & \textbf{Size (MB)} & \textbf{Accuracy (\%) }  \\
%          \hline
%          \multirow{2}{*}{RF} & Baseline & 2.086 & 40.51 & 92.50 \\
%          & \system & 1.815 & 61.46 & 90.33 \\
%          \hline
%          \multirow{2}{*}{LR} & Baseline & 0.094 & 36.83 & 96.13 \\
%          & \system & 0.373 & 58.56 & 92.94 \\
%          \hline
%          \multirow{2}{*}{DNN} & Baseline & 0.137 & 36.83 & 94.74 \\
%          & \system & 0.384 & 58.59 & 92.90 \\
%     \end{tabular}
%     \label{tab:system_performance}
% \end{table}

\subsection{Overhead of Proposed System}
We also examine the overhead of inference with \system with the goal that \system should not exert significant additional overhead compared to the baseline method of using the service model directly. The results are shown in Table \ref{tab:system_performance} for the HAR dataset. 

While there is a minor loss in accuracy across service model types, the inclusion of an autoencoder and adversarial detection layer in the inference pipeline leads to an approximate 50\% increase in the overall application size. However, a significant increase in storage comes from the \emph{scipy} library used for computation of leakage rate components. With additional mathematical optimizations in the implementation, the dependency on this library can be removed, reducing the application size further.

We also note the $\sim 1.4$-fold and $\sim 1.1$-fold times increase in inference runtime for the LR and DNN models respectively. However, it is important to note that the tradeoff between runtime and protection of proprietary information may not negatively impact the user experience given the generally low inference times ($<$ 0.3 seconds). 
%By implementing more efficient algorithms, and enabling memory management and resource utilization via lower level languages, it is possible to further decrease latency. 
Conversely, there is a $\sim 0.16$-fold decrease in query runtime for the RF model. We hypothesize this is due to the RF's sensitivity towards the number of input features; since the autoencoder converts the input into a lower dimensional vector, the RF has to compute fewer operations, resulting in improved runtime efficiency. For all model types, the adversarial detector ran concurrently with the inference service, averaging to 0.158 seconds. 

\textbf{Key Takeaway: } While prioritizing higher service provider privacy adversely affects storage, runtime, and accuracy, this tradeoff can be deemed insignificant in practice. By enabling the detector to run in the background for the previous timestep, the service latency is unaffected by the detector.

\section{Related Work}
% \begin{table*}[t!]
%     \centering
%     \small
%     \begin{tabular}{c|lcccc}
%         \textbf{Scope} & \textbf{Method} & \textbf{Edge} & \textbf{Impact } & \textbf{High Dimensional} & \textbf{Non-Linear} \\
%          &  & \textbf{Friendly} & \textbf{Benign Clients} & \textbf{Friendly} & \textbf{Space Friendly} \\
%         \hline 
%         \multirow{4}{*}{Input} & Quantifying feature space explored \cite{kesarwani2018model} & \cmark & & & \\
%          & Evaluating closeness to decision boundaries \cite{quiring2018forgotten} & \cmark & & & \\
%          & Adaptively misinforming for OOD queries \cite{kariyappa2020defending} & \\
%          & Analyzing query distribution \cite{grosse2017statistical, meng2017magnet, juuti2019prada} \\
%          \hline
%          \multirow{2}{*}{Model} & Protecting parameters with differential privacy \cite{tramer2016stealing} & \\
%          & Training diverse ensemble models \cite{kariyappa2021protecting} & \\
%          \hline
%          Output & Modifying prediction probabilities \cite{tramer2016stealing, lee2019defending, orekondy2019prediction} & \cmark & & & \\
%     \end{tabular}
%     \caption{Summary of existing defenses against model extraction attacks.}
%     \label{tab:extraction-solutions}
% \end{table*}

%attack space
Model extraction attacks aim to learn information about the model itself. They use a trained ML model to extract model parameters and learn an equivalent shadow model to poison with adversarial examples \cite{papernot2017practical, oh2019towards} or monetize off of the model in ML as a service applications \cite{tramer2016stealing}. 
Since training data is often unavailable, extraction attacks employ querying techniques to exploit the model. Beyond random querying methods, more sophisticated methods have been used where datasets from similar domains have been considered \cite{correia2018copycat, orekondy2019knockoff} or synthetic queries have been objectively crafted \cite{papernot2017practical, juuti2019prada, kariyappa2021maze}.
However, most applications of model extraction have been motivated by user privacy. That is, the privacy issue has revolved around learning sensitive information from the training data. Closer to our work, prior works have examined model extraction attacks in security applications such as spam or fraud detection \cite{ateniese2015hacking}. Here the goal was to evade detection using the extracted model. We explore a more generic approach for on-device deployment.

%solution space
Towards preventing such attacks, prior works have proposed methods to detect and prevent adversarial usage. %proposed into various methods from modifying prediction probabilities to detecting adversarial queries. 
We present these in Table \ref{tab:extraction-solutions}.
Papernot et al. employed two existing methods in adversarial ML, namely adversarial training and defensive distillation for DNNs \cite{papernot2017practical}. Tramer et al. discussed rounding confidence scores and using differential privacy on training points or model parameters as potential defense mechanisms \cite{tramer2016stealing}. Kariyappa et al. proposed training an ensemble of diverse models with a diversity objective that requires models in the ensemble to produce dissimilar predictions \cite{kariyappa2021protecting}. However, these protection methods are not fully generalizable across model types, primarily being applicable to DNN-based models only.

Juuti et al. proposed the first generic technique to detect model extraction attacks using analysis of successive queries by checking for deviations from normal distribution \cite{juuti2019prada}. However, they assume on-device deployment relies on platform security mechanisms to provide localized isolation preventing white-box attacks. Furthermore, their detection process starts when a client queries at least 100 samples. 
%Furthermore, they assume the adversary has architectural knowledge about the model with the black-box access. 
Our work proposes an end-to-end solution that effectively thwarts extraction attacks in both white-box and black-box scenarios. Furthermore, our approach promptly identifies adversarial behavior from the initiation of usage, ensuring early detection and prevention.

Other works have proposed perturbing OOD queries with adaptive misinformation \cite{kariyappa2020defending} or modifying prediction probabilities \cite{tramer2016stealing, orekondy2019prediction}. However, these works assume adversarial queries are OOD in nature. Furthermore, modifying predictions based on usage may not be applicable in situations where maintaining application performance for benign users is a critical requirement.

% Prior works have also proposed distributed learning methods for training models on the device \cite{xie2019slsgd, atrey2021preserving} and explored methods to do efficient inference \cite{lee2019device}. 
% Our work focuses on not only detecting adversarial usage in a generalizable manner but protecting against it in an end-to-end fashion.

% \vspace{-1em}
\section{Conclusion}
In this work, we examined the implication of leaking proprietary information in ML models deployed on user devices through querying attacks. Our results demonstrated that such attacks can be used to recover up to 100\% of the output space and exploit decision boundaries with a 100\% success rate. We proposed an end-to-end framework, \system, that supports the deployment of ML-based applications on user devices. In \system, we introduced an adversarial detection layer that leverages an autoencoder model to classify usage type as benign or adversarial. Our evaluation of \system, conducted on two widely-used datasets, demonstrates its ability to detect adversarial usage with an 89\% accuracy within a small number of queries.

Although \system has exhibited effectiveness against simpler attacks targeting on-device models, we acknowledge the need for future research to explore more intricate attack scenarios. Additionally, examining the effect of colluding adversaries and expanding the attacks to generative AI models, which are inherently different from classification models examined in this paper, are interesting scopes for future work.

%%
%% The acknowledgments section is defined using the "acks" environment
%% (and NOT an unnumbered section). This ensures the proper
%% identification of the section in the article metadata, and the
%% consistent spelling of the heading.
\begin{acks}
We thank the anonymous reviewers for their helpful comments. We also thank Meet Vadera, Priyanka Mammen, Bhawana Chhaglani, and Phuthipong Bovornkeeratiroj for their feedback. This research was supported in part by NSF grants 2211302, 2211888, 2213636,  2105494, 1908536, Army Research Lab contract W911NF-17-2-0196 and Adobe. Any opinions, findings, and conclusions, or recommendations expressed in this material are those of the authors and do not necessarily reflect the views of the funding agencies.
\end{acks}

\balance
%%
%% The next two lines define the bibliography style to be used, and
%% the bibliography file.
\bibliographystyle{ACM-Reference-Format}
\bibliography{ref}

\end{document}